\documentclass{article} 
\usepackage{iclr2021_conference,times}

\usepackage{amsmath,amsfonts,bm}

\def\eqref#1{equation~\ref{#1}}

\def\1{\bm{1}}

\DeclareMathAlphabet{\mathsfit}{\encodingdefault}{\sfdefault}{m}{sl}
\SetMathAlphabet{\mathsfit}{bold}{\encodingdefault}{\sfdefault}{bx}{n}

\usepackage{hyperref}
\usepackage{url}
\usepackage{multirow}
\usepackage{times}
\usepackage{enumitem}
\usepackage{latexsym}
\usepackage{graphicx}
\usepackage{subcaption}
\usepackage{csquotes}
\usepackage{physics}
\usepackage{textcomp}
\usepackage{amsmath}
\usepackage{amssymb}
\usepackage{nicefrac}       
\usepackage{booktabs}       
\usepackage{lipsum}         
\usepackage{xargs}          
\usepackage{amsfonts}       
\usepackage{array}

\title{Learning to Summarize Long Texts with Memory Compression and Transfer}

\author{Jaehong Park\thanks{$\;$ Authors contributed equally to this work}~~$^{1}$, Jonathan Pilault$^{*1,2,3}$, and Christopher Pal$^{1,2,3,4,5}$ \\
  $^1$Element AI \\ $^2$Mila, $^3$Polytechnique Montreal, $^4$University of Montreal, 
  $^5$Canada CIFAR AI Chair \\
  \texttt{firstname.lastname@elementai.com} }

\iclrfinalcopy 
\begin{document}

\maketitle

\begin{abstract}
We introduce \emph{Mem2Mem}, a memory-to-memory mechanism for hierarchical recurrent neural network based encoder decoder architectures and we explore its use for abstractive document summarization. Mem2Mem transfers ``memories" via readable/writable external memory modules that augment both the encoder and decoder. Our memory regularization compresses an encoded input article into a more compact set of sentence representations. Most importantly, the memory compression step performs implicit extraction without labels, sidestepping issues with suboptimal ground-truth data and exposure bias of hybrid extractive-abstractive summarization techniques. By allowing the decoder to read/write over the encoded input memory, the model learns to read salient information about the input article while keeping track of what has been generated. 
Our Mem2Mem approach yields results that are competitive with state of the art transformer based summarization methods, but with \emph{16 times fewer parameters}.
\end{abstract}

\section{Introduction}
Automatic summarization is the automated process of reducing the size of an input text while preserving its most relevant information content and its core semantics. Techniques for summarization are often characterized as being either: \emph{Extractive} or \emph{Abstractive}. \emph{Extractive} methods construct summaries by combining the most salient passages (usually whole sentences) of a source text; a process similar to human's way of identifying the right information. One way to achieve extractive summarization is to define the problem as a sentence classification task, using some form of  representation of the sentences in a document \citep{nallapati2016classify, cheng2016neural}. To avoid content overlap issues, previous work has used sentence reranking \citep{rank_extractSumm} or sentence ordering by extracting sentences recurrently \citep{chen-bansal-2018-fast}. \emph{Abstractive} methods generate summaries by generating new sentence constructs ``from scratch'', or from representation of document content, a process that is conceptually more similar to the notion of paraphrasing. Abstractive text summarization has attracted interest since it is capable of generating novel formulations of summaries using language generation models conditioned on the source text. Several attention-based Recurrent Neural Network (RNN) encoder-decoders have been introduced to tackle varying text generation issues of standalone abstractive sequence-to-sequence (seq2seq) models. Copy and pointer mechanisms \citep{gu-etal-2016-incorporating, tu-etal-2016-modeling, vinyals2015pointer, asee}, for example, have enabled decoders to better generate unseen words, out-of-vocabulary words and named entities.
 
Most recently, hybrid extractive and abstractive architectures have been proposed and have shown promising results in both quantitative performance measures and human evaluations. In such set-ups, the extractive model first selects salient sentences from a source article, and the abstractive model paraphrases the extracted sentences into a final summary. The majority of current state-of-the-art abstractive summarization models\footnote{Excluding summarization models using large scale pre-trained language models such as BERT \citep{devlin2018bert}} are based on the hybrid approach \citep{chen-bansal-2018-fast, gehrmann-etal-2018-bottom, hsu-etal-2018-unified, bae2019summary, tlm2019extractive}. 
Nonetheless, hybrid models can be limited by three disadvantages. First, since ground-truth labels for extractive summarization are usually not provided, extractive labels must be generated by a potentially suboptimal algorithm \citep{hsu-etal-2018-unified, tlm2019extractive}. The performance of models trained with such labels is therefore bounded by the quality of the performance of the extractive heuristics. Second, since ground-truth binary labels for recurrently extracted sentences are typically teacher forced as in \citet{chen-bansal-2018-fast}, ``exposure bias'' \citep{exposurebias2015} may negatively affect content selection performance at inference. Finally, given that the hard extraction step is not differentiable, existing hybrid models typically require multi-step training (not end-to-end) \citep{gehrmann-etal-2018-bottom, tlm2019extractive} or reinforcement learning \citep{chen-bansal-2018-fast, bae2019summary} to train the whole model.

In this paper, we introduce a novel abstractive summarization model that incorporates an intermediate extractive step but does not require labels for this type of extractive content selection, and it is fully end-to-end trainable. To achieve this, we propose a new memory augmented encoder-decoder (MAED) architecture \citep{wang-etal-2016-memory, yogatama2018memory, taxonomy_memnet, v_mem_encdec} called Mem2Mem. Mem2Mem has 2 memorization modes: (1) absorb key information of the encoded source sequence via a compression mechanism, and (2) sequentially update the external memory during target summary generation. Without using extractive ground-truth labels, we find in our analysis that Mem2Mem's compression mechanism behaves as an implicit sentence extractor that stores sentence representations of the salient content. The choice of sentence representations is only guided by the memory regularization and conditional language modeling loss of the decoder, thus avoiding exposure bias from maximizing the likelihood of sequential binary extraction labels. Finally, the encoded memory is transferred to the decoder memory, which is iteratively refined during the decoding process. To our knowledge, Mem2MeM is the first abstractive summarization model that uses memory compression for sentence extraction and that directly employs the memorized representations during summary generation. We empirically demonstrate the merits of this approach by setting a new state-of-the-art on long text abstractive summarization tasks on the Pubmed, arXiv and Newsroom datasets \citep{cohan-etal-2018-discourse,grusky2018newsroom}. Our contributions are three fold:
\begin{enumerate}[noitemsep]
    \item We introduce the Mem2Mem approach that (i) stores salient sentence-level representations via memory compression, (ii) transfers memory from encoder to decoder and (iii) updates the memory as the summary generation proceeds.
    \item Unlike previous works, our model combines the best of extractive and abstractive summarization in a fully end-to-end trainable manner without supervision on the extraction step.
    \item Our method yields results that are competitive with state of the art transformer \citep{vaswani2017attention} language model based techniques, but with substantially fewer parameters (only 6\% of the parameters used by a recent state of the art transformer based method).
\end{enumerate}

\begin{figure*}[t]
    \center{\includegraphics[scale=0.19] 
    {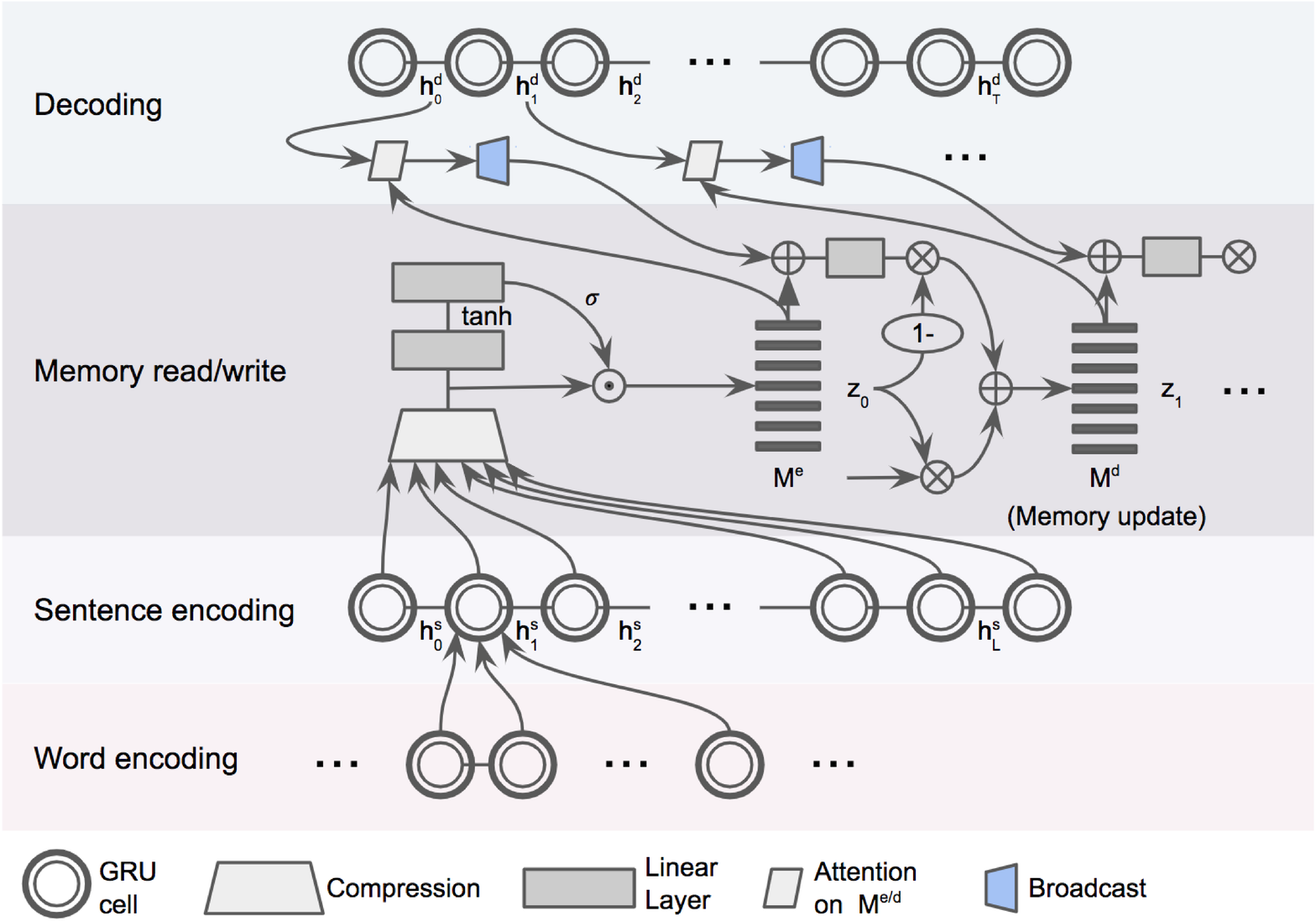}}
    \caption{\label{fig:model} \small Mem2Mem architecture: Sentence-level representations from the document encoder are reduced to a fix sized encoder memory \(\mathbf{M}_{E}\). The encoder memory is transferred to the decoder memory \(\mathbf{M}_{D}\) and is read using a RAM like mechanism. The resulting memory readout vector is used to condition sentence and word-level attention. The decoder hidden states and the memory readout vector are then used to update the memory state \(\mathbf{M}_{D}\) via a gated write operation ($z_i$).}
\end{figure*}

\section{Background}
\label{sec_background}
For our baseline, we use a hierarchical recurrent encoder-decoder structure (HRED) based seq2seq model from \citet{nallapati-etal-2016-abstractive} and \citet{cohan-etal-2018-discourse}. 
The HRED has two encoder GRUs: a sentence encoder and a document encoder. Given an input sentence of length \(N\), the sentence encoder takes a sequence of token embeddings \(\mathbf{x}\) and transforms it into a sequence of hidden states.
\begin{equation}
\label{eq_sent_rnn}
    \mathbf{h}^{(w)}_{1}, 
    \ldots, \mathbf{h}^{(w)}_{N} = \textrm{GRU}_{sen}(\mathbf{x}_{1}, \mathbf{x}_{2}, \ldots, \mathbf{x}_{N})
\end{equation}
The last hidden state of the sentence encoder is used as a corresponding sentence embedding \(\mathbf{s}\). 
The sequence of sentence embeddings \(\mathbf{s}\) are then processed by the document encoder.
\begin{equation}
\label{eq_doc_rnn}
    \mathbf{h}^{(s)}_{1}, 
    \ldots, \mathbf{h}^{(s)}_{L} = \textrm{GRU}_{doc}(\mathbf{s}_{1}, \mathbf{s}_{2}, 
    \ldots, \mathbf{s}_{L})
\end{equation}
where \(\mathbf{s}_{j}\) is the \(j\) th sentence embedding, \(\mathbf{h}^{(s)}_{j}\) is the associated document encoder hidden state and \(L\) is the total number of sentences in the document.

The decoder GRU generates a target summary one token at a time. At each decoding step \(t\), the decoder creates a decoder hidden state \(\mathbf{h}^{(d)}_{t}\). The decoder then obtains a context vector \(\mathbf{c}_t\) via $\gamma_{ti}$, an alignment between \(\mathbf{h}^{(w)}_{i}\) and \(\mathbf{h}^{(d)}_{t}\). $\gamma_{ti}$ is computed by combining token level attention $\alpha$ and sentence level attention $\beta$, such that:
\begin{equation}
\label{eq_word_attn}
    \alpha_{ti} = \textrm{Attn}(\mathbf{h}^{(d)}_{t}, {\mathbf{h}^{(w)}_{i}}), \;\;\;\;\;\;\;\;
    \beta_{tj} = \textrm{Attn}(\mathbf{h}^{(d)}_{t}, \mathbf{h}^{(s)}_{j}), \;\;\;\;\;\;\;\;
    \gamma_{ti} = \frac{\beta_{tm(i)}\alpha_{ti}}{\sum_{l=1}^{N_{d}} \beta_{tm(l)}\alpha_{tl}}
\end{equation}
where \(m(l)\) denotes the index of the sentence corresponding to the \(l\) th word, and \(N_d\) is the total number of tokens in the input document. \textrm{Attn} in equation (\ref{eq_word_attn}) 
is defined as in \citet{luong2015effective}.

The probability distribution of the next target word \(y_{t}\) is estimated using the decoder hidden state \(\mathbf{h}^{(d)}_{t}\) and the context vector \(\mathbf{c}_t\). The training objective \(\mathcal{L}\) is the average negative log likelihood of the target word \(y_t\) over the whole ground truth summary of length \(T\).
Finally, the pointer generator and decoder coverage method in \citet{asee} are used for the baseline HRED. More details of the baseline architecture can be found in Appendix 1.

\section{Memory-to-Memory Mechanism (Mem2Mem)}
Mem2Mem has three main features. (1) An encoder memory bank that compresses a large set of encoder hidden representations into a smaller subset of salient representations. (2) Read/write operations that allow the decoder to read and update the encoder memory bank with new information during summary generation. (3) Token generation that is conditioned on the extracted sentence representations accessed from the dynamic memory. The Mem2Mem components can be seamlessly integrated with existing HRED architectures. In essence, the whole process can be seen as \textit{extraction} followed by \textit{generation}. The architecture is depicted in Figure \ref{fig:model}.

\subsection{Memory Compression on Encoder} \label{sec:encmem}
The aim of having an external memory on the encoder is to create a fixed size representation that reduces the set of $L$ hidden representations from $\textrm{GRU}_{doc}$ to a subset of $r$ representations. From a sequence of sentence-level document encoder hidden representations \(\mathbf{h}^{(s)}_{1}, \mathbf{h}^{(s)}_{1}, \ldots \mathbf{h}^{(s)}_{L}\), we construct an intermediate 2-D matrix \(\mathbf{H} \in \mathbb{R}^{L \times d}\), where \(d\) is the document encoder hidden size,
\begin{equation}
\label{eq_sent_emb}
    \mathbf{H} =
    \begin{bmatrix}
    {\mathbf{h}^{(s)}_{1}} \;
    {\mathbf{h}^{(s)}_{2}} \;
    \ldots \;
    {\mathbf{h}^{(s)}_{L}}
    \end{bmatrix}^T.
\end{equation}
A smaller sized memory bank is generated by taking a linear combination of each row in \(\mathbf{H}\). The weight vector for the linear combination \(\mathbf{a}\) is computed with self-attention mechanism.
\begin{equation}
\label{eq_self_attn_weights}
    \mathbf{a} = \textrm{softmax} (w_{a1}^{T} \textrm{tanh} (\mathbf{W}_{a2} \mathbf{H}^{\top}))
\end{equation}
where \(w_{a1} \in \mathbb{R}^{d_a}\), \(\mathbf{W}_{a2} \in \mathbb{R}^{d_a \times d}\),  and \(d_a\) is the size of the hidden layer which is a hyperparameter. To capture various aspects of the input document, \(\mathbf{a}\) is extended to a multi-head memory write attention matrix \(\mathbf{A}\) with \(r\) heads,
$    \mathbf{A}= \textrm{softmax} (\mathbf{W}_{a1} \textrm{tanh} (\mathbf{W}_{a2} \mathbf{H}^{\top}))
$
where \(\mathbf{W}_{a1} \in \mathbb{R}^{r \times d_a}\) and \(r\) is a hyperparameter. This results in \(r\) different convex combinations of \(\mathbf{H}\), which gives us the final multi-head encoder memory matrix \(\mathbf{M}_{E} \in \mathbb{R}^{r \times d}\),
$    \mathbf{M}_{E} = \mathbf{AH}.
$
%
%
To ensure the attention weights in various heads focus on a diverse set of salient input sentences, we propose a novel regularization loss for memory compression. The following regularization term encourages the diversity of compressed encoded states.
\begin{equation}
\label{eq_enc_memreg}
    \mathcal{L}^{(comp)} = {||(\mathbf{AA}^{\top} - \mathbf{I})||_F}^2
\end{equation}
where $||\bullet||_F$ is the Frobenius norm. The regularization loss \(\mathcal{L}^{(comp)}\) achieves two goals simultaneously: (1) It promotes diversity over the \(r\) sentence representations stored in the memory, thereby reducing the risk of redundant information. (2) It hardens the attention probabilities of each head, assuring that each memory slot is approximately associated to a single sentence representation. As a result, the encoder memory \(\mathbf{M}_{E}\) essentially performs implicit extraction over the encoder hidden states in a fully differentiable manner. Figure~\ref{fig_enc_mem_attn_analysis} in Appendix shows the effect of regularization on the encoder memory compression. Note that no supervision exists on this extractive step and the memory compression is only guided by the memory regularization and back-propagated error signals from the target summary generation.

\subsection{Read/Write operations on Decoder} \label{sec:decmem}
Once the encoder memory is constructed, the context read from the memory is used to augment both attention mechanism and the target token generation.
As a first step, the encoder memory \(\mathbf{M}_{E}\) is transferred to the decoder and used as an initial state of the decoder memory \(\mathbf{M}_{D}\). At every time step \(t\), the decoder reads from the memory and generates a memory read vector \(\mathbf{m}_t\). Specifically, the decoder takes the weighted sum over \(r\) memory slots via a RAM like attention mechanism:
\begin{equation}
    \psi_{tk} = \textrm{Attn}(\mathbf{h}^{(d)}_{t}, \mathbf{M}_{D}(k)),
\;\;\;\;\;\;\;\;
    \mathbf{m}_t = \sum_{k=1}^{r} \psi_{tk} \mathbf{M}_{D}(k),
    \label{eq_mem_attn}
\end{equation}
where \(\mathbf{M}_{D}(k)\) is the vector representation of the \(k\) th head or slot of \(\mathbf{M}_{D}\). \(\mathbf{h}^{(d)}_{t}\) is then combined with \(\mathbf{m}_t\) and generate a memory augmented decoder hidden state 
%
$    \mathbf{h}^{(m)}_{t} = \mathbf{W}_m [\mathbf{h}^{(d)}_{t}; \mathbf{m}_t]
$
and the next token estimation of the baseline system is replaced with \(\mathbf{h}^{(m)}_{t}\). 
As a consequence, the attention over the source text and the prediction of the target token are conditioned on the memory read \(\mathbf{m}_t\). Thus, there is a direct link between the contents of the memory and the text generation.

During the summary generation process, the semantics of the source sequence that is kept in the decoder memory needs to be modified. The memory write operation of Mem2Mem enables the memory to log the history of what has been attended and generated. The decoder memory write operation outlined below removes and adds information using a gated mechanism to forget used memories and update each memory slot. The gating mechanism is conditioned on the the memory content \(\mathbf{M}_{D}\), the memory read vector \(\mathbf{m}_t\), and the decoder hidden state \(\mathbf{h}^{(d)}_{t+1}\):

\begin{equation}
\label{eq_mem_write_1}
    \mathbf{z}^{k}_{t} = \boldsymbol{\sigma} (\mathbf{W}_{z_1}\mathbf{h}^{(d)}_{t+1} + \mathbf{W}_{z_2} \mathbf{m}_{t} + \mathbf{W}_{z_3} \mathbf{M}_{D}(k))
\end{equation}
\begin{equation}
\label{eq_mem_write_2}
    \mathbf{u}^{k}_t = \tanh (\mathbf{W}_{u_1}\mathbf{h}^{(d)}_{t+1} + \mathbf{W}_{u_2} \mathbf{m}_{t} + \mathbf{W}_{u_3} \mathbf{M}_{D}(k))
\end{equation}
\begin{equation}
\label{eq_mem_write_3}
    \mathbf{M}_{D}(k) := \mathbf{z}^{k}_{t} \odot \mathbf{M}_{D}(k) + (1 - \mathbf{z}^{k}_{t}) \odot \mathbf{u}^{k}_{t}
\end{equation}

Although text generation is directly conditioned on the memory context \(\mathbf{m}_t\), the benefit of having memory was limited in our preliminary experiments. We observed only a few memory slots were repetitively attended during decoding.
To ensure that the Mem2Mem decoder fully utilizes its memory to improve generation, we propose another regularization term \(\mathcal{L}^{(read)}\).
\begin{equation}
\label{eq_dec_memreg_1}
    \mathbf{c}^{(m)}_t = \sum_{k=1}^{r} \psi_{tk} \mathbf{M}_{E}(k),
\;\;\;\;\;\;\;
    \mathbf{c}^{(s)}_t = \sum_{j=1}^{L_s} \beta_{tj} \mathbf{h}^{(s)}_j,
\;\;\;\;\;\;\;
    \mathcal{L}^{(read)} = \frac{1}{T} \sum_{t=1}^{T} {||\mathbf{c}^{(m)}_t - \mathbf{c}^{(s)}_t||_2}
\end{equation}
where \(\mathbf{M}_{E}(k)\) is the vector representation of the \(k^{th}\)  memory head of \(\mathbf{M}_{E}\) and \(\mathbf{c}^{(s)}_t\) is a sentence level context vector. The regularization assumes that if the memory context \(\mathbf{m}_t\), which is a sentence-level representation, is combined properly within the decoding, the sentence level context vector \(\mathbf{c}^{(s)}_t\) would correlate with it. The initial state of the decoder memory \(\mathbf{M}_{E}\) is used to compute \(\mathbf{c}^{(m)}_t\) since the representation of \(\mathbf{M}_{D}\) deviates from the original representation space of \(\mathbf{h}^{(s)}_j\) due to the write operation. The final training objective of Mem2Mem becomes as follows.
\begin{equation}
\label{eq_train_loss_final}
    \mathcal{L} + \lambda_1 \mathcal{L}^{(comp)} + \lambda_2 \mathcal{L}^{(read)}
\end{equation}
where the weights for regularization \(\lambda_1\) and \(\lambda_2\) are hyperparameters.


\begin{table*}[t]
\caption{Results on the PubMed, arXiv and Newsroom dataset. Each type corresponds to purely abstractive (A), extractive (E), or extractive-abstractive hybrid (H) approaches. TLM uses the GPT-2 model \citep{radford2019gpt2} that has 16 times larger parameters than Mem2Mem. The highest ROUGE scores for abstractive methods are boldfaced. All ROUGE scores have a 95\% confidence interval of at most ±0.25 as reported by the official ROUGE. Results taken from: $^{1}$\citet{tlm2019extractive} and $^{2}$\citet{cohan-etal-2018-discourse}. * Newsroom abstractive summarization test set results.}
\begin{center}
\footnotesize
\begin{tabular}{|c|c|c|ccc|c}
	\hline 
		\multirow{2}*{Model} & \multirow{2}*{Type} & \# of
		& \multicolumn{3}{c|}{ROUGE-1/2/L}  \\
        & & params. & PubMed & arXiv & Newsroom* \\ \hline
        Lead-10 / 3 \scriptsize{(Newsroom)}$^{1}$ & E & - & \footnotesize{37.45/14.19/34.07} & \footnotesize{35.52/10.33/31.44} & \footnotesize{13.7/2.4/11.2} \\
        Sent-CLF$^{1}$ & E & - & \footnotesize{45.01/19.91/41.16} & \footnotesize{34.01/8.71/30.31} & \footnotesize{15.4/2.7/12.8} \\
        Sent-PTR$^{1}$ & E & - & \footnotesize{43.30/17.92/39.47} & \footnotesize{42.32/15.63/38.06} & \footnotesize{15.9/2.8/13.0} \\
        \hline
        Attn-Seq2Seq$^{1,2}$ & A & - & \footnotesize{31.55/8.52/27.38} & \footnotesize{29.3/6.00/25.56} & \footnotesize{6.21/1.07/5.68} \\
        Ptr-Gen-Seq2Seq$^{1,2}$ & A & - & \footnotesize{35.86/10.22/29.69} & \footnotesize{32.06/9.04/25.16} & \footnotesize{14.66/2.26/11.44} \\
        Discourse-aware$^{2}$ & A & 14.3M & \footnotesize{38.93/15.37/35.21} & \footnotesize{35.80/11.05/31.80} & \footnotesize{-/-/-} \\
        TLM$^{1}$ & A & 234M & \footnotesize{37.06/11.69/34.27} & \footnotesize{39.65/12.15/35.76} & \footnotesize{\textbf{20.40}/\textbf{6.90}/\textbf{17.10}} \\
        TLM$^{1}$ & H & 234M & \footnotesize{\textbf{42.13}/16.27/\textbf{39.21}} & \footnotesize{41.62/14.69/\textbf{38.03}} & \footnotesize{20.10/6.50/16.60} \\
        Mem2Mem \footnotesize{(ours)} & A & 14.7M & \footnotesize{42.06/\textbf{16.56}/38.11} & \footnotesize{\textbf{41.81}/\textbf{14.99}/37.24} & \footnotesize{20.00/6.60/16.80} \\
    \hline
\end{tabular}
\end{center}
\label{tab_rouge}
\end{table*}

\section{Related Work} \label{section:related_work}
Recent works in abstractive summarization have leveraged intermediate content selection. In these approaches, writing a summary is factorized into two steps: \textit{extraction} and \textit{generation}. An extractor is used to prioritize and select the most important part of the input text. The extractor is normally trained on a sequence of binary labels where each label indicates whether the corresponding text unit should be selected or not. The level of extraction can be word-level \citep{gehrmann-etal-2018-bottom,cho2019mixture} or sentence-level \citep{chen-bansal-2018-fast,hsu-etal-2018-unified,bae2019summary,tlm2019extractive}. As the ground truth for extraction is typically missing, heuristics that measure n-gram overlap with the target summary are used to build extractive oracles. Similar to other approaches, Mem2Mem performs sentence-level extraction to deal with long source articles\footnote{In preliminary experiments, we applied word-level selection on the PubMed and arXiv datasets, which led to poor results}. Mem2Mem determines the alignment between source and target sentences in a latent space without relying on possibly suboptimal extractive heuristics. In addition, sentence extraction is not sequentially done in Mem2Mem, which addresses the exposure bias issue \citep{exposurebias2015}.
\raggedbottom \newline \par
Memory Augmented Encoder Decoder (MAED) architectures \citep{wang-etal-2016-memory, yogatama2018memory, taxonomy_memnet, v_mem_encdec} have been proposed for conditional natural language generation tasks, such as machine translation \citep{Kaiser_mem_nmt} and image captioning \citep{Park_mem_captioning}. Using differentiable read and write operations to an external module, MAED can represent non-local context of RNNs with enhanced memory capacity. Such models are able to store temporally distant information of large input sequences, a feature that is particularly useful for long text summarization. In the context of short text abstractive summarization, \citet{kim2018abstractive} proposed a memory architecture named multi-level memory networks (MMN). MMN can flexibly reduce representations at different levels of document hierarchy into a fixed size external memory. Authors used multi-layer dilated Convolutional Neural Networks (CNN) \citep{yu2015multi,oord2016wavenet} to build a hierarchical representation of the document. Mem2Mem also constructs memory from the hierarchical representation of the document, but by compressing it into a sparse set of sentence representations. Further, MMN's memory representations remain static throughout the decoding process while Mem2Mem dynamically updates its memory, which is more effective in learning long term dependency. Lastly but not least, our work proposes novel regularization for memory read and compression.

\section{Results and Discussion} \label{sec:results}
\subsection{Experiment Setup}
We evaluate Mem2Mem on the PubMed, arXiv \citep{cohan-etal-2018-discourse} and Newsroom abstractive datasets \citep{grusky2018newsroom} which are large scale summarization datasets. The average lengths of source articles and target summaries are 3016/203 (PubMed), 4938/220 (arXiv), and 751/30 (Newsroom) respectively. They are up to 6 times longer than the widely used CNN/DailyMail dataset (781/56) \citep{hermann2015teaching,asee}. Our pre-processing and training setups are identical to \citet{cohan-etal-2018-discourse} and \citet{tlm2019extractive}. More details on training and evaluation can be found in Appendix 2. For quantitative evaluation, we use the ROUGE metric \citep{lin2004looking} and report F-1 ROUGE scores.

\subsection{Results}
Table~\ref{tab_rouge} shows the ROUGE scores on three summarization datasets. The hybrid type (H) refers to models that use two-step extractive-abstractive summarization. 
On the PubMed dataset, the TLM model shows the highest scores in R-1 and R-L. Mem2Mem is close to those scores and shows higher R-2 (+0.29) scores with 16 times less parameters ($14.7$M vs. $234$M). It achieves such performance over the TLM model without ground truth labels for sentence extraction. We also reiterate that Mem2Mem is trained completely end-to-end whereas the hybrid TLM requires separate training for the extractor and the conditional transformer language model. We find similar results on the arXiv and Newsroom datasets. On the arXiv dataset, Mem2Mem even surpasses the transformer based TLM model in R-1 (+0.19) and R-2 (+0.3) scores. Mem2Mem also shows competitive results on the Newsroom abstractive dataset.

\begin{table}[t]
\caption{Model ablation study on the PubMed dataset.}
\begin{center}
\small
\begin{tabular}{|l|ccc|}
	\hline 
		\multirow{2}*{Model} & 
		\multicolumn{3}{c|}{ROUGE} \\
        & 1 & 2 & L  \\ \hline
        Baseline HRED & 40.02 & 15.82 & 36.28 \\
        \quad + Encoder Mem & 40.62 & 15.89 & 36.85 \\
        \quad + Decoder Mem & 40.51 & 15.9 & 36.64 \\
        \quad + Mem Transfer & 41.27 & 16.24 & 37.38 \\
        \quad + Reg \(\mathcal{L}^{(comp)}\) & 41.82 & 16.51 & 37.76 \\
        \quad + Reg \(\mathcal{L}^{(read)}\) & \textbf{42.06} & \textbf{16.56} & \textbf{38.11} \\
    \hline
\end{tabular}
\end{center}
\label{tab_pubmed_ablation}
\end{table}

\begin{table}[t]
\caption{ROUGE scores of unsupervised extractive methods on the PubMed and the arXiv dataset. The result of baseline extractive methods is from \cite{tlm2019extractive}.}
\begin{center}
\small
\begin{tabular}{|c|c|ccc|}
	\hline 
		\multirow{2}*{Data} & \multirow{2}*{Model}
		& \multicolumn{3}{c|}{ROUGE}  \\
        & & 1 & 2 & L  \\ \hline
        \multirow{4}*{PubMed} & Lead-10  & 37.45 & 14.19 & 34.07 \\
        & LexRank & 39.19 & 13.89 & 34.59 \\
        & Mem2Mem & \textbf{41.77} & \textbf{16.20} & \textbf{37.81} \\
        \cline{2-5}
        & Gold Ext & 47.76 & 20.36 & 39.19 \\ \hline\hline
        \multirow{4}*{arXiv} & Lead-10 & 35.52 & 10.33 & 31.44 \\
        & LexRank & 33.85 & 10.73 & 28.99 \\
        & Mem2Mem & \textbf{41.76} & \textbf{14.95} & \textbf{37.22} \\
        \cline{2-5}
        & Gold Ext & 44.25 & 18.17 & 35.33 \\ \hline
\end{tabular}
\end{center}
\label{tab_mem_extraction}
\end{table}

\subsection{Ablation Study}
To assess the importance of Mem2Mem components, we conduct an ablation study on the PubMed dataset. 
Table~\ref{tab_pubmed_ablation} shows the effects of adding different Mem2Mem add-ons on ROUGE scores. \textit{+Encoder Mem} is the baseline HRED augmented with the encoder memory described in section~\ref{sec:encmem}. The result demonstrates that memory context indeed enhances the performance measures of the generated summaries. \textit{+Decoder Mem} adds the write operation to the memory but without memory transfer. In this case, the decoder memory \(\mathbf{M}_D\) is initialized with zeros not with the encoder memory \(\mathbf{M}_E\).
Compared to the Baseline HRED, the result shows that the write mechanism on the decoder memory helps the generation even without the memory transfer. This indicates that the summary writing process largely benefits from accessing long term contextual information of the output text. 

Transferring memory (+\textit{Mem Transfer}) brings substantial improvements in ROUGE scores. It seems to be crucial to initiate the summary generation from the selected memory representations, showing the importance of the memory compression and transfer steps.
Furthermore, it can be observed that discouraging redundancies over the encoder memory head via \(\mathcal{L}^{(comp)}\) leads to additional improvements on all ROUGE scores. Finally, adding another regularization \(\mathcal{L}^{(read)}\) on the decoder memory read operation completes the Mem2Mem architecture and achieves the best ROUGE scores.

\subsection{Implicit Extraction via Memory Compression}
Our initial hypothesis was that the encoder memory \(\mathbf{M}_E\) would pick a set of the most salient input sentences for summarization. To confirm, we analyze the quality of the extractive summarization by memory compression. Concretely, we concatenate the sentences with the highest attention weight of each memory head to generate a summary.
Table~\ref{tab_mem_extraction} shows the ROUGE scores of different unsupervised extractive summarization methods on the PubMed and arXiv datasets. The extractive summarization performed by the Mem2Mem's memory compression outperforms existing unsupervised extractive summarization baselines. The result indicates that Mem2Mem's memory compression is able to prioritize amongst a large set of input sentences without ground truth sentence labels.

\begin{figure*}
\begin{subfigure}[h]{0.5\linewidth}
\includegraphics[width=\linewidth]{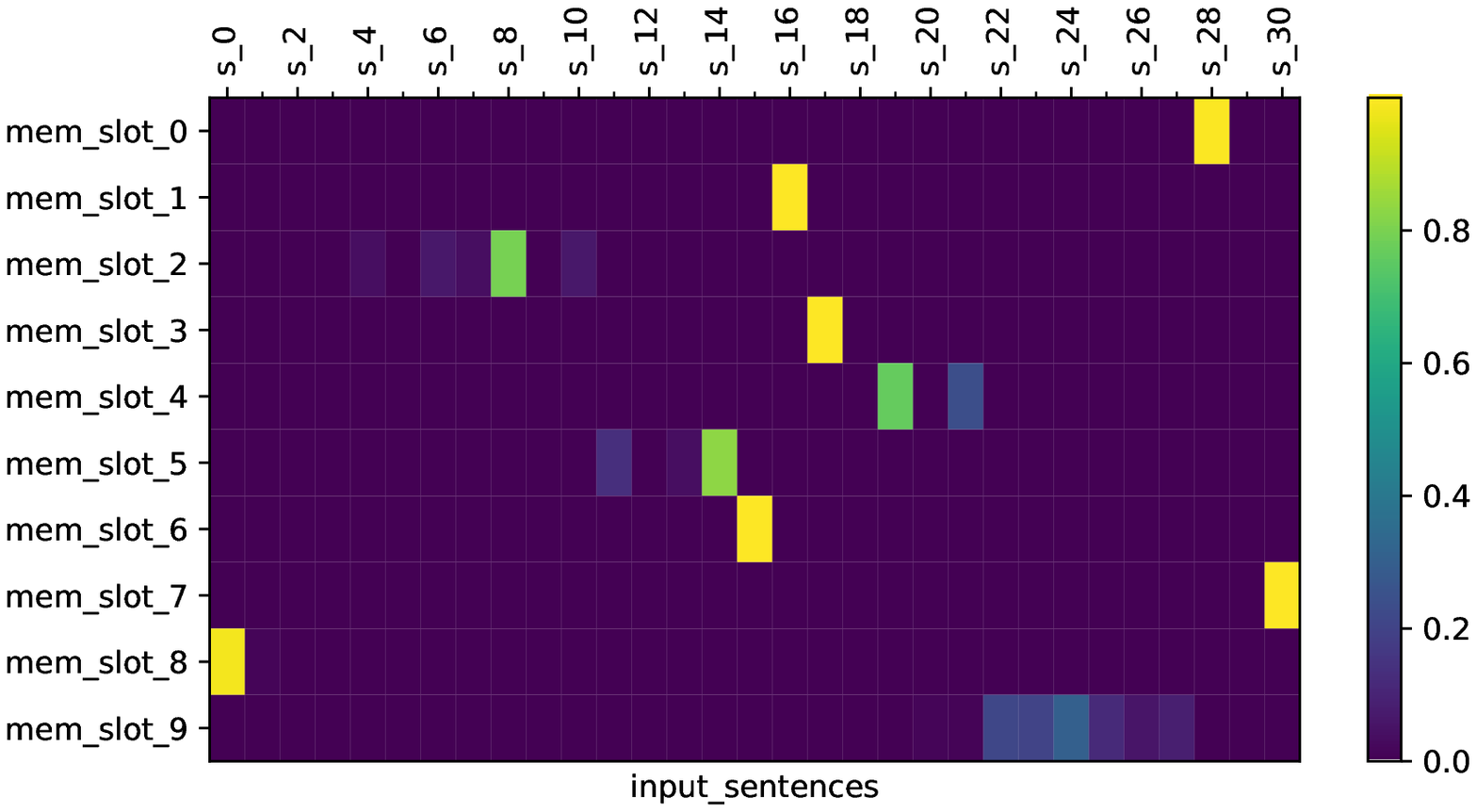}
\caption{Multi-head memory write attention matrix \(\mathbf{A}\)}
\end{subfigure}
\hfill
\begin{subfigure}[h]{0.5\linewidth}
\includegraphics[width=\linewidth]{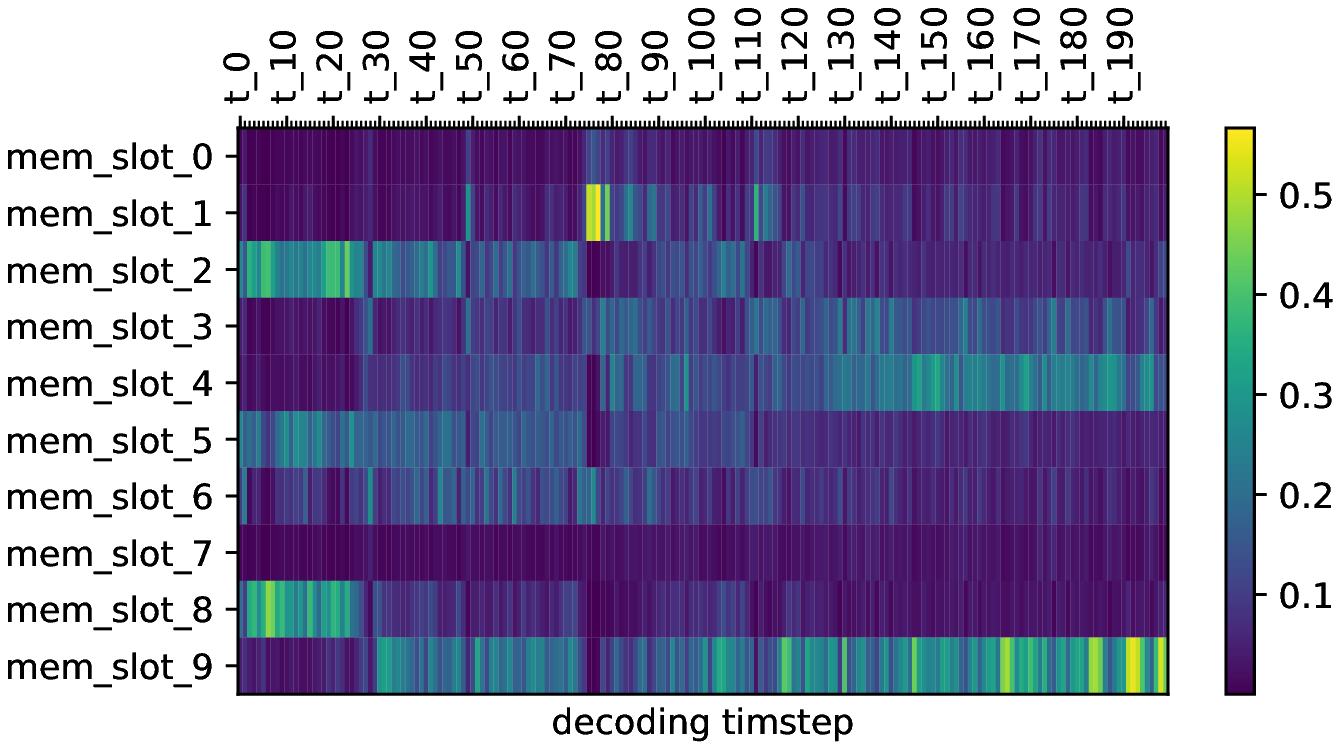}
\caption{Memory read attention matrix \(\mathbf{\Psi}\)}
\end{subfigure}
\caption{Multi-head memory write attention matrix \(\mathbf{A}\) and memory read attention matrix \(\mathbf{\Psi}\). In (a), rows denote memory heads or slots and columns indicate input sentence indices. In (b), columns indicate decoding time steps.}
\label{fig_dec_mem_attn_analysis}
\end{figure*}

\subsection{Dynamic Memory Read by Decoder}
The benefit of implicit extraction is maximized when the extracted representations are properly consumed in the text generation. 
To understand the link between the representations stored in the memory and the summary generation, we analyze the memory read attention weights \(\psi_{tk}\) in equation~(\ref{eq_mem_attn}) throughout the decoding. Figure~\ref{fig_dec_mem_attn_analysis} (b) shows that the Mem2Mem decoder fully utilizes all memory representations. We also find a pattern that memory read attention weights are mostly concentrated on the front part of the source text in the beginning of the decoding (mem slots 2,5,8), and gradually moves to the latter part (mem slots 3,4,9). This demonstrates Mem2Mem's ability to update the read operation to dynamically capture relevant input contexts during the summary generation.

\subsection{Abstractiveness of the Summary}
To analyze the abstractiveness of generated summaries, we present the ratio of output summary n-grams present in the original input article. Table~\ref{tab_n_gram} shows that Mem2Mem copies less n-grams than the baseline HRED. Compared to the baseline HRED, Mem2Mem generates approximately 10\% more novel 5-grams and 7\% more 10-grams respectively. The result along with higher ROUGE scores highlights the Mem2Mem's ability to generate novel words for abstractive summarization while staying focused on important parts of the article. Although TLM \citep{tlm2019extractive} \footnote{The authors  provided example summaries from their model.} shows the highest abstractiveness, Mem2Mem achieves its result with a significanly smaller model.

\begin{table}[t]
\caption{Percentage ratios of output summary n-grams found in the PubMed original input article.}
\begin{center}
\small
\begin{tabular}{|c|cccc|}
	\hline 
	    \multirow{2}*{Models} & \multicolumn{4}{c|}{n-grams} \\
		 & 5 & 10 & 15 & 20 \\ \hline
		Reference & 12.4 & 3.2 & 1.3 & 0.9 \\
		Baseline HRED & 38.1 & 26.0 & 17.4 & 13.4 \\
		TLM-hybrid & 24.7 & 10.8 & 7.0 & 4.1 \\
		Mem2Mem (ours) & 28.3 & 19.2 & 17.6 & 10.7 \\
	\hline
\end{tabular}
\end{center}
\label{tab_n_gram}
\end{table}

\begin{table}[t]
\caption{Results of human evaluation. Maximum score for each criterion is 5.}
\begin{center}
\small
\begin{tabular}{|c|cccc|}
	\hline 
	    \multirow{2}*{Models} & \multicolumn{4}{c|}{Human evaluation scores} \\
		 & COH & INF & REL & FLU \\ \hline
		Baseline HRED & 3.02 & 2.98 & 2.85 & 3.08 \\
		TLM-hybrid & \textbf{3.67} & 3.11 & 3.43 & \textbf{3.51} \\
		Mem2Mem (ours) & 3.59 & \textbf{3.13} & \textbf{3.46} & 3.38 \\
	\hline
\end{tabular}
\end{center}
\label{tab:human_eval}
\end{table}

\subsection{Human Evaluation}
We also perform a human evaluation to assess the quality of generated summaries. For the human evaluation, 20 random arXiv testset article-summary pairs are presented to three Amazon Mechanical Turk workers. Workers judge generated summaries on four different aspects: Coherence (COH: does the summary make sense), Informativeness (INF: are the most important points of the article captured), Redundancy (RED: does the summary repeat itself), and Fluency (FLU: how fluent is the summary). Table~\ref{tab:human_eval} shows that Mem2Mem obtains the highest scores in Informativeness and Redundancy. Its Coherence score is also close to the best score by TLM. Fluency is an advantage for the transformer-based TLM as expected but Mem2Mem is close and still greatly superior to vanilla hierarchical encoder decoders. While the ROUGE difference between the baseline and Mem2Mem are substantial in Table~\ref{tab_pubmed_ablation}, the results of the human evaluation show more pronounced differences in the quality of generated text. For output summary examples, please refer to Table ~\ref{tab:qualitative} in Appendix 4.


\section{Conclusion}
This work proposes Mem2Mem, a novel MAED based mechanism for very long text abstractive summarization. Mem2Mem involves two memory types: A static encoder memory for compressing input texts and a dynamic decoder memory which refines the generation process. Memory transfer between them links two memories and maximizes the benefit of content extraction aimed for summarization. Different from existing hybrid extractive and abstractive approaches, Mem2Mem incorporates an extraction step without ground truth sentence labels and multi-step training. We demonstrate the effectiveness of Mem2Mem by showing promising results on the PubMed, arXiv, and Newsroom summarization datasets with an order of magnitude less parameters than competing transformer-based models. The Mem2Mem's memory compression can be generalized to other domains that require text generation guided by content selection. In future work, we will extend and validate the strength of our approach on a variety of language learning tasks.

\bibliography{iclr2021_conference}
\bibliographystyle{iclr2021_conference}

\newpage
\appendix
\section{Appendix}

\subsection{Baseline Architecture}
In addition to the hierarchical recurrent encoder-decoder (HRED) architecture described in section~\ref{sec_background}, the following features are used for both baseline and Mem2Mem architectures.

\subsubsection{Pointer Generator network}
In order to handle out-of-vocabulary (OOV) token predictions, pointer generator in \cite{asee} is used to copy words directly from the input document. At each step \(t\), the decoder decides whether to predict the next target word from the input text or the generation mechanism. The pointer generator computes \(z_t\) which denotes the probability of choosing \(P_{g}\) for sampling the next target token.
\begin{equation}
\label{eq_p_z}
    z_t = \textrm{softmax}(w^{T}_{c} \mathbf{c}_t + w^{T}_{d} \mathbf{h}^{(d)}_{t} + w^{T}_{x} \mathbf{x}^{'}_{t})
\end{equation}
where \(\mathbf{x}^{'}_{t}\) is the embedding of the previous target token.
The probability \(z_t\) is used as a variable for soft switch between generating a word from the vocabulary (\(P_g\)) or directly copying from the source document (\(P_c\)). The probability of copying a word \(w\) from the source text is calculated based on the attention weights \(\gamma\).
\begin{equation}
\label{eq_p_copy}
    P_{c}(y_t=w | y_{1:t-1}) = \sum_{i:x_i=w}\gamma_{ti}
\end{equation}

Note that \(P_{c} (y_t=w|y_{1:t-1}) = 0\) if \(w\) does not exist in the source document. Likewise, \(P_{g}(y_t=w | y_{1:t-1}) = 0\) if \(w\) is an out of vocabulary word.
Combining two probability distributions, the final probability of the next word \(y_t\) being \(w\) is as follows.
\begin{equation}
\label{eq_p_final}
    \begin{aligned}
    P(y_t=w | y_{1:t-1}) = z_t P_{g}(y_t=w|y_{1:t-1}) + (1-z_t) P_{c}(y_t=w|y_{1:t-1})
    \end{aligned}
\end{equation}

\subsubsection{Decoder coverage}
It is well known that RNN sequence-to-sequence models tend to suffer from repeated phrases when generating long target sequences. \cite{asee} tackled this issue by keeping track of the attention coverage. More concretely, the coverage vector \(\mathbf{cov}_t\) at the decoding step \(t\) is computed by taking the summation of the token-level attention weights  \(\alpha\) until the last step \(t-1\).
\begin{equation}
\label{eq_cov}
    \mathbf{cov}_t = \sum_{t'=0}^{t-1} \alpha_{t'}
\end{equation}

To inform the decoder of the history of attention weights, the coverage vector is fed into the token-level attention mechanism, which modifies the equation (\ref{eq_word_attn}) to the following equation.
\begin{equation}
\label{eq_attn_cov}
    \begin{aligned}
    \alpha_{ti} = \textrm{softmax} \Big( v^{\top} \tanh (W_{e}\mathbf{h}^{(w)}_{i} + W_{d}\mathbf{h}^{(d)}_{t} + w_{c}~\mathbf{cov}_{t}^{T}) \Big)
    \end{aligned}
\end{equation}

\subsection{Training Details}
We generally follow the pre-processing steps in \cite{cohan-etal-2018-discourse} for the PubMed and arXiv datasets. The maximum number of sections is set to 4 and the maximum number of tokens for each section is 500. The length of the target summary is limited to 200 tokens. 

Single-layer bidirectional GRUs \citep{cho2014learning} are used for the sentence and the document encoders. The decoder is also a single layer GRU. All GRUs have the hidden size of 256. The dimensionality of token embeddings is 128 and embeddings are trained from scratch. The vocabulary size is limited to 50,000. Batch size is 16 and Adam \citep{kingma2014adam} with learning rate 2\(e^{-4}\) is used for training. Maximum gradient norm is set to 2. We train all models for 15 epochs. At the test time, beam search with the beam size 4 is used for decoding. 

For Mem2Mem hyperparameters, the number of heads for the memory compression is 10 and the self-attention hidden size is 128. The weights \(\lambda_1\) and \(\lambda_2\) for the regularization \(\mathcal{L}^{(comp)}\) and \(\mathcal{L}^{(read)}\) are 0.0001 and 0.01 respectively.

\clearpage
\subsection{The effect of regularization}
The following figures show the effect of regularization in Mem2Mem.

\begin{figure*}[h]
\begin{subfigure}[h]{0.5\linewidth}
\includegraphics[width=\linewidth]{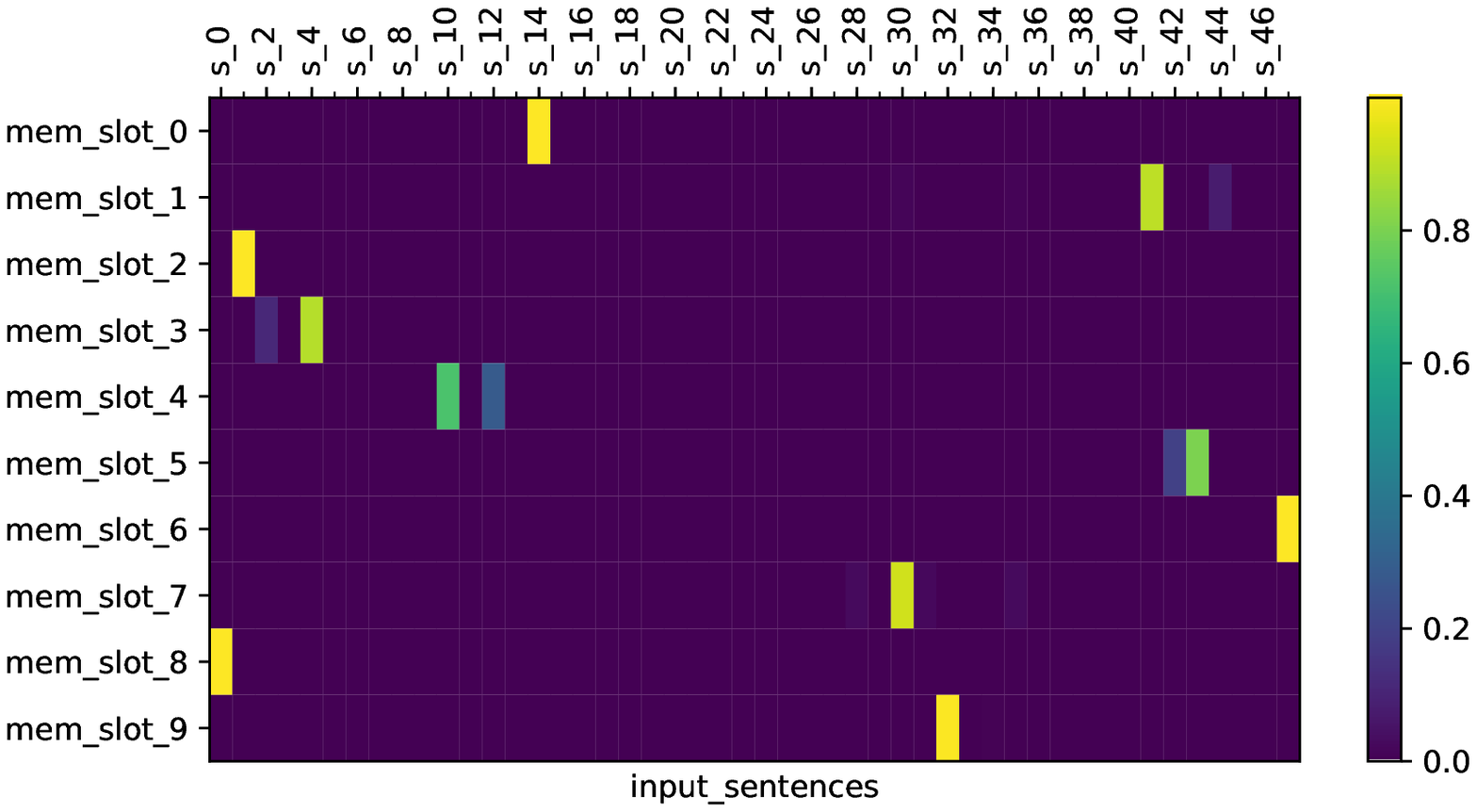}
\caption{With regularization \(\mathcal{L}^{(comp)}\)}
\end{subfigure}
\hfill
\begin{subfigure}[h]{0.5\linewidth}
\includegraphics[width=\linewidth]{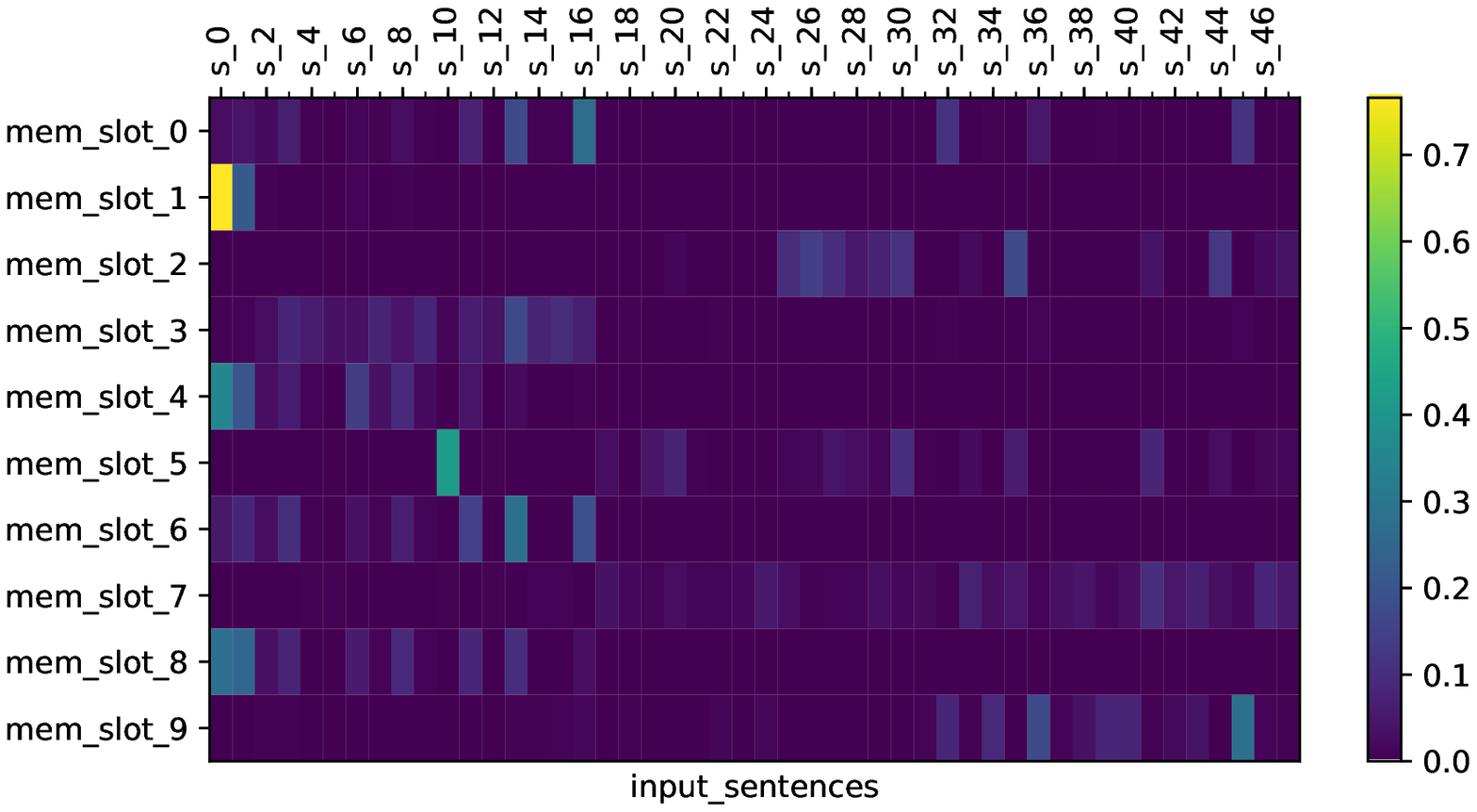}
\caption{Without regularization \(\mathcal{L}^{(comp)}\)}
\end{subfigure}
\caption{The effect of regularization on memory compression. Examples of the multi-head encoder memory write attention matrix \(\mathbf{A}\) are illustrated. Rows denote memory heads or slots and columns indicate input sentence indices. Note that the regularization loss \(\mathcal{L}^{(comp)}\) removes the redundancy over different memory heads and guides each slot to focus on a single sentence.}
\label{fig_enc_mem_attn_analysis}
\end{figure*}

\begin{figure*}[h]
\begin{subfigure}[h]{0.5\linewidth}
\includegraphics[width=\linewidth]{figure/analysis_attn.eps}
\caption{With regularization \(\mathcal{L}^{(read)}\)}
\end{subfigure}
\hfill
\begin{subfigure}[h]{0.5\linewidth}
\includegraphics[width=\linewidth]{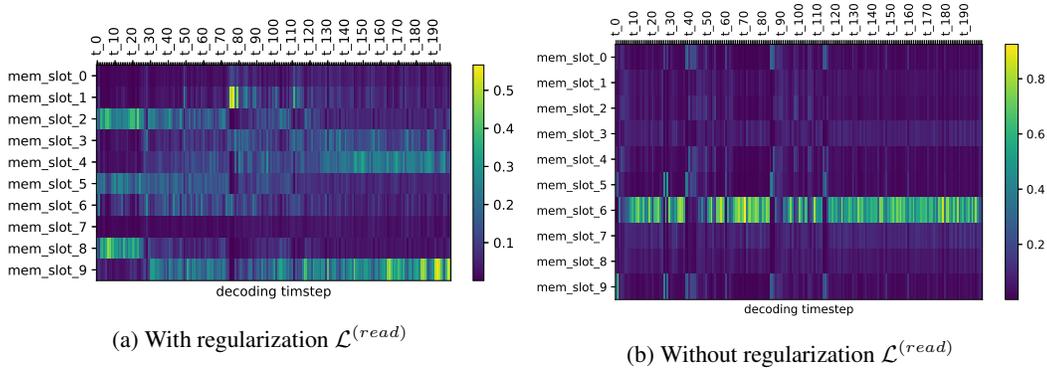}
\caption{Without regularization \(\mathcal{L}^{(read)}\)}
\end{subfigure}
\caption{The effect of regularization on memory read. Examples of the decoder memory read attention matrix \(\mathbf{\Psi}\) are illustrated. Rows denote memory heads or slots and columns indicate decoding time steps. Note that the regularization loss \(\mathcal{L}^{(read)}\) encourages the model to fully utilize the compressed memory representations.}
\label{fig_dec_mem_attn_analysis_appendix}
\end{figure*}

\subsection{Qualitative results}
Reference and system generated summaries from the test set of the arXiv dataset are shown in the following table. Compared to the baseline, Mem2Mem shows much less repetition and produces more concise summaries. Even the TLM model repeats several phrases or sentences.

\begin{table}[t]
\caption{Reference and system generated summaries from the test set of the arXiv dataset.}
\begin{center}
\footnotesize
\begin{tabular}{|m{13.5cm}|}
\hline
\textbf{Reference} --- we study the behavior of simple principal pivoting methods for the p-matrix linear complementarity problem ( p - lcp ) . we solve an open problem of morris by showing that murty s least index pivot rule ( under any fixed index order ) leads to a quadratic number of iterations on morris s highly cyclic p - lcp examples . we then show that on k - matrix lcp instances , \_ all \_ pivot rules require only a linear number of iterations. as the main tool , we employ \_ unique sink orientations \_ of cubes , a useful combinatorial abstraction of the p - lcp . \\
\hline
\textbf{Baseline HRED} --- the third author of this paper still vividly recollects his visit to victor klee at the university of washington (seattle) in august 2000 . the third author of this paper still vividly recollects his visit to victor klee at the university of washington (seattle) in august 2000 . the third author of this paper still vividly recollects his visit to victor klee at the university of washington (seattle) in august 2000 . in this paper . , we introduce the digraph model behind p - lcps and show that the simplex method is polynomial - time for which the number of iterations is \_ exponential \_ in the number of variables and constraints . \\
\hline
\textbf{TLM-hybrid} --- we study linear complementarity problems with sufficient matrices and the criss - cross method . we use a novel notion of \_ linear complementarity \_ to derive linear complementarity problems with sufficient matrices and the ``linear'' version of the ``quadratic'' version of the ``linear'' version of the ``linear'' version . this yields a new family of np-complete problems for which both linear complementarity and the ``quadratic'' version of the ``linear'' version of the ``quadratic'' version are possible . \\
\hline
\textbf{Mem2Mem} --- in this paper we introduce a digraph model behind p - lcps and show that the simplex method can be used to solve a linear program ( lp ) for which the number of iterations is \_ exponential \_ in the number of variables and constraints on the number of iterations . we show that the number of arithmetic operations does not depend on the bit lengths of the input numbers . we also show that the number of iterations required to solve the linear complementarity problem can be used to determine the expected performance of the lp . \\
\hline
\hline
\textbf{Reference} --- we have developed a procedure for the classification of eclipsing binaries from their light - curve parameters and spectral type. the procedure was tested on more than 1000 systems with known classification, and its efficiency was estimated for every evolutionary status we use. the procedure was applied to about 4700 binaries with no classification, and the vast majority of them was classified successfully . systems of relatively rare evolutionary classes were detected in that process, as well as systems with unusual and/or contradictory parameters. also, for 50 previously unclassified cluster binaries evolutionary classes were identified. these stars can serve as tracers for age and distance estimation of their parent stellar systems. the procedure proved itself as fast, flexible and effective enough to be applied to large ground based and space born surveys, containing tens of thousands of eclipsing binaries.
 \\
\hline
\textbf{Baseline HRED} --- in this paper we present a novel procedure for determination of the evolutionary status of eclipsing binaries. the procedure is based on the most comprehensive set of rules for the classification of eclipsing binaries, which is the world s principal database of eclipsing binary systems with available classification. the main goal of this paper is to develop a fast and effective procedure for determination of the binary s spectral type or color index. the procedure is based on the most comprehensive set of rules for the classification of eclipsing binaries, which is the world s principal database of eclipsing binary systems with available classification. the main goal of this paper is to develop a fast and effective procedure for determination of the evolutionary status of eclipsing binaries. the method is based on a catalogue which includes a number and the eclipsing variables, and a classification procedure. we find a good fit of all known parameters, which can not have any value in a large range. we find a significant fraction (mp, depth) for each of these systems, and that a given number is @xmath0 mag
 \\
\hline
\textbf{TLM-hybrid} --- a fast and effective procedure for determination of the evolutionary status of eclipsing binaries is presented. a complete set of rules is developed that utilizes all available data for a specific class of eclipsing binaries , while requiring only light curve parameters and an estimate of the binary s spectral type or color index. the procedure is tested with the catalogue of eclipsing binaries ( cev ), which is the world s principal database of eclipsing binaries with available classification. the main goal of our work is to develop a fast and effective procedure for determination of the evolutionary status of eclipsing binaries. we apply it to cev binaries with already available classification. the main objective of this work is to develop a fast and effective procedure for determination of the evolutionary status of eclipsing binaries. the main objective of our work is to develop a fast and effective procedure for determination of the evolutionary status of eclipsing binaries. we apply it to cev binaries with already available classification. the main objective of this work is to demonstrate the usefulness of our method, and to propose a possible application of it to cev binaries with already known classification.
 \\
\hline
\textbf{Mem2Mem} --- in this paper we present a novel procedure for determination of the evolutionary status of eclipsing binaries, while requiring only light curve parameters and an estimate of the binary s spectral type or color index. the procedure is based on the most comprehensive set of rules for the classification of eclipsing binaries, which can be used to quickly characterize large numbers of eclipsing binaries. the procedure is tested with the catalogue of eclipsing variables ( dr \& sh ), which is the world s principal database of eclipsing binary systems with available classification. we find that the number of fully characterized eclipsing binaries can be used to quickly characterize large numbers of eclipsing binaries.
 \\
\hline
\end{tabular}
\end{center}
\label{tab:qualitative}
\end{table}
\end{document}